\documentclass[letterpaper,10pt,conference]{ieeeconf}
\IEEEoverridecommandlockouts
\usepackage{graphicx}
\usepackage{stfloats}
\usepackage{float}
\usepackage{placeins}
\usepackage{booktabs}
\usepackage{colortbl}
\usepackage{xcolor}
\definecolor{tablebest}{RGB}{120,160,220}
\definecolor{tablesecond}{RGB}{210,228,245}
\newcommand{\bestcell}[1]{\cellcolor{tablebest}\textbf{#1}}
\newcommand{\secondcell}[1]{\cellcolor{tablesecond}#1}
\usepackage{multirow}
\usepackage{wrapfig}
\usepackage{algorithm}
\usepackage{algorithmicx}
\usepackage{algpseudocode}
\usepackage{url}
\usepackage{caption}
\usepackage{amsmath}
\usepackage{amssymb}
\usepackage{amsfonts}
\makeatletter
\let\NAT@parse\undefined
\makeatother
\usepackage[colorlinks=true,linkcolor=blue,citecolor=blue,urlcolor=blue]{hyperref}
\usepackage[all]{hypcap}
\newcommand{\figref}[1]{\hyperref[#1]{Fig.~\ref*{#1}}}
\newcommand{\tabref}[1]{\hyperref[#1]{Table~\ref*{#1}}}
\newcommand{\tabrefs}[2]{\hyperref[#1]{Tables~\ref*{#1}} and~\hyperref[#2]{\ref*{#2}}}
\newcommand{\secref}[1]{\hyperref[#1]{Section~\ref*{#1}}}
\title{\LARGE \bf
DRIVE-Nav: Directional Reasoning, Inspection, and Verification for Efficient Open-Vocabulary Navigation
}
\author{Maoguo Gao$^{1}$,\hspace{0.7em} Zejun Zhu$^{2}$,\hspace{0.7em} Zhiming Sun$^{2}$,\hspace{0.7em} Zhengwei Ma$^{1}$,\hspace{0.7em} Longze Yuan$^{1}$\\
Zhongjing Ma$^{1,3}$,\hspace{0.7em} Zhigang Gao$^{1,3}$,\hspace{0.7em} Jinhui Zhang$^{1,3}$,\hspace{0.7em} and Suli Zou$^{1,3,*}$%
\thanks{*Corresponding author. Email: {\tt\small sulizou@bit.edu.cn}}%
\thanks{$^{1}$Beijing Institute of Technology}%
\thanks{$^{2}$DeepBlue College}%
\thanks{$^{3}$The National Key Laboratory of Autonomous Intelligent Unmanned Systems (KAIUS), School of Automation}%
}

\begin{document}
\maketitle
\vspace{-1.2em}
\thispagestyle{empty}
\pagestyle{empty}
\begin{abstract}
Open-Vocabulary Object Navigation (OVON) requires an embodied agent to locate a language-specified target in unknown environments.  Many zero-shot methods  rely on frontier-candidate reasoning under incomplete observations,  while topology-aware methods reduce candidate redundancy but may still introduce panoramic inspection overhead and repeated reconsideration. We present DRIVE-Nav, a structured framework that organizes exploration around persistent directions rather than raw frontiers. By inspecting encountered directions more completely and restricting subsequent decisions to still-relevant directions within a forward 240\textdegree{} view range, DRIVE-Nav reduces redundant revisits and improves path efficiency. The framework extracts and tracks directional candidates from weighted Fast Marching Method (FMM) paths, maintains representative views for semantic inspection, and combines vision--language-guided prompt enrichment with cross-frame verification to improve grounding reliability. Experiments on HM3D-OVON, HM3Dv1, HM3Dv2, and MP3D demonstrate strong overall performance and consistent efficiency gains. On HM3D-OVON, DRIVE-Nav achieves 50.2\% SR and 32.6\% SPL, improving the previous best method by 1.9\% SR and 5.6\% SPL. It also delivers the best SPL on HM3Dv1, HM3Dv2, and MP3D and transfers to a physical humanoid robot. Real-world deployment also demonstrates its effectiveness. \href{https://coolmaoguo.github.io/drive-nav-page/}{Project page}
\end{abstract}
\noindent\textbf{Keywords}: Open-Vocabulary Object Navigation, Embodied Navigation, Mobile Robotics.
\section{Introduction}

\begin{figure}[!t]
\centering
\includegraphics[width=0.88\columnwidth,keepaspectratio]{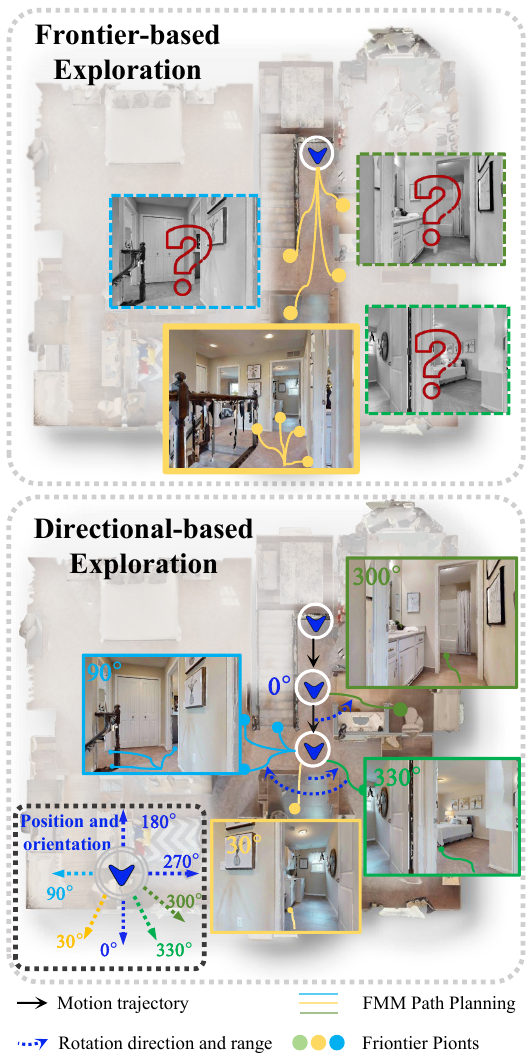}
\caption{Comparison of exploration paradigms. Frontier-based exploration selects among frontier points without  observing the unexplored regions beyond them . Directional-based exploration  instead groups candidate directions around the agent, where each inset image denotes the representative view observed from the agent’s position along the corresponding direction.}
\label{fig:framework}
\end{figure}

Robust navigation is a fundamental capability for autonomous robots operating in
real-world environments~\cite{batra2020objectnav}. In particular,
object-goal navigation requires an agent to reach a target specified by its
semantic category in previously unseen scenes. The problem becomes even more
challenging in open-vocabulary settings, where agents must generalize to a wide
and unconstrained range of object categories while exploring unfamiliar
environments~\cite{40}. Conventional learning-based navigation methods often
rely on extensive training data and struggle to generalize to new scenes~\cite{5,19,15}.
Recent advances in large pre-trained models, including large language models
and vision--language models, provide new opportunities for zero-shot navigation
by enabling agents to leverage semantic knowledge to infer likely object
locations and guide exploration~\cite{7,32,28,27}.

Current  zero-shot exploration methods expose different decision interfaces.
Frontier-candidate methods select discrete points along the boundary of
explored free space~\cite{32,28,27,yamauchi1997frontier}. For these interfaces,
multiple candidates may correspond to the same navigable direction, making
model-side  decisions redundant. As illustrated in the  frontier-based exploration
panel of \figref{fig:framework},  the four frontier points can lead to the same
navigable path, so choosing among them at the current position provides little
meaningful distinction~\cite{28,27}.
 In addition, point-level candidates provide only indirect evidence . The robot can
observe the explored side of  a frontier, while the semantic content beyond the
branch remains unseen before route commitment. In \figref{fig:framework}, for example, frontier-based selection observes only the current forward view, leaving the side branches visually unknown, whereas direction-based exploration inspects newly revealed branch directions before selecting the next direction. Even when candidates are
augmented with surrounding objects~\cite{28},  frontier-side visual evidence~\cite{27},
or  scene graphs~\cite{26} , the model  still infers branch utility from geometric
proxies rather than directly  maintained branch-level evidence . In contrast,
direction-based exploration moves the agent to a more informative decision
position and performs sufficient rotation to inspect the outgoing directions
before committing to the next route.

 Topology-aware methods mitigate these frontier-level limitations. In particular,
VoroNav~\cite{wu2024voronav} constructs a Reduced Voronoi Graph, stops at
informative Voronoi junction nodes, rotates to collect surrounding observations,
and selects among topological branches. This already moves beyond raw frontier
points and validates the value of junction-level decision making. However, the
observation policy is still tied to a full 360$^\circ$  look-around when the
agent reaches an RVG node or mid-term goal, after which the current panoramic
images are matched to neighboring nodes for farsight descriptions. Thus, even
when only a subset of outgoing directions requires new visual evidence, the
agent still pays a fixed panoramic-inspection cost at that decision location.
As illustrated in the bottom panel of \figref{fig:framework}, an agent
may rotate 360$^\circ$ at both junctions , although only 60$^\circ$ is needed at
the first junction, and at the second junction only 30$^\circ$ to the left
followed by 120$^\circ$ to the right.

 Studies on human wayfinding suggest that intersections are critical decision
points that prompt information seeking and spatial decision making before route
commitment~\cite{brunye2018spatial}. Inspired by this perception--decision
coupling , DRIVE-Nav organizes exploration around  persistent directions rather
than transient frontier candidates or node-local branch observations. Each
direction is a stable group of frontiers sharing a similar local bearing, where
the bearing is induced by the
weighted FMM path from the robot to each frontier. Once a direction is selected,
the agent advances along its farthest frontier until the evolving frontier
structure exposes new forward exits at a more informative decision position.
DRIVE-Nav then inspects these newly revealed forward directions within the
240$^\circ$ range by rotating only over the angular span needed to face them,
rather than executing a fixed full-circle scan. The rear-facing range mainly
corresponds to the path already traversed, so it is not treated as a new
forward exploration choice. The representative views captured at these
decision-relevant positions provide richer branch-level visual evidence for
Qwen3-VL~\cite{bai2025qwen3} to select the next  direction, judge target visibility, and generate
target-aware  prompt refinements for SAM3~\cite{carion2025sam} , followed by
cross-frame verification in a unified semantic loop.

Beyond exploration, open-vocabulary grounding remains vulnerable to false
positives caused by distractors, partial views, and semantic ambiguity. A
candidate that appears plausible in a single frame may not correspond to the
true target once additional views are observed. We therefore further introduce
a cross-frame Qwen3-VL~\cite{bai2025qwen3} verification module that cross-checks each detected
candidate against recent multi-view observations before writing it to the
object map, so that only consistently confirmed targets are accepted and
spurious detections are suppressed early.

Experiments on HM3D-OVON~\cite{40}, HM3Dv1~\cite{ramakrishnan2021hm3d}, HM3Dv2~\cite{yadav2023habitat}, and
MP3D~\cite{chang2017matterport3d} demonstrate that DRIVE-Nav
consistently improves both success rate and path efficiency over strong
baselines. We further deploy DRIVE-Nav on a physical humanoid robot, confirming
its real-world feasibility.

In summary, this work makes the following contributions:
\begin{itemize}
    \item A  direction-centric exploration interface for zero-shot OVON  that
    replaces redundant point-level  frontier decisions with temporally
    persistent, motion-grounded direction entities, and acquires
    representative views for newly revealed forward directions.

    \item A unified semantic loop  that uses direction-level  representative
    views as shared evidence for exploration and target grounding,  enabling
    Qwen3-VL to select directions, assess target visibility, and generate
    inspection-guided  SAM3 prompt refinements for weakly grounded targets.

    \item A cross-frame Qwen3-VL verification mechanism that couples temporal
    semantic confirmation with navigation, improving target-grounding
    reliability under open-vocabulary ambiguity.

    \item Comprehensive validation on HM3D-OVON,
    HM3Dv1, HM3Dv2, and MP3D,  showing strong performance and consistent
    path-efficiency gains across diverse benchmarks.
\end{itemize}

\section{Related work}

\subsection{Zero-Shot Object Navigation}
Zero-Shot Object Navigation (ZSON) aims to locate language-specified target objects in unseen environments without task-specific training and to generalize to novel object categories.
Early methods improve generalization through learned multimodal embeddings, modular exploration policies, or semantic association priors ~\cite{5,19,15,guo2024actpept}, but still rely on substantial task-specific training or learned priors.
Recent studies  reduce reliance on task-specific training by combining foundation models with explicit spatial or topological representations for semantic exploration~\cite{7,32,28,27,wu2024voronav,26,zhang2025apexnav}.
Building on this trend, DRIVE-Nav further improves zero-shot  exploration by organizing frontier candidates into persistent directions and using representative views for inspection and grounding verification.

\subsection{Scene Representation and Exploration Strategy}
Training-free zero-shot navigation methods differ mainly in how they represent the observed scene and convert it into exploration goals. L3MVN~\cite{28}  builds a semantic map and uses an LLM to identify target-related frontiers. VLFM~\cite{27} , InstructNav~\cite{14}, and ApexNav~\cite{zhang2025apexnav} rank frontier or map candidates with vision-language value maps  or target-centric semantic scores. Scene-graph methods such as SG-Nav~\cite{26}, UniGoal~\cite{yin2025unigoal}, and MSGNav~\cite{huang2025msgnav} encode object and spatial relations for LLM-based reasoning, while VoroNav~\cite{wu2024voronav} uses a topology-aware  representation to select among Voronoi branches.

 The frontier-oriented methods above enrich frontier selection with semantic maps, value maps, or scene-graph cues, but still choose from frontier or map candidates. As discussed in the Introduction, this candidate-level interface can be redundant and provides only indirect evidence about unexplored branches. VoroNav~\cite{wu2024voronav} moves the decision interface to topological branches, but still relies on full panoramic scans at junctions and may reconsider rearward directions along the already traversed path. DRIVE-Nav instead converts frontier candidates into persistent directions and inspects newly observed forward directions within the necessary angular range, enabling direction-level exploration without repeated full 360$^\circ$ scans.

\subsection{Target Identification for ZSON}
Reliable target identification is essential for zero-shot  ObjectNav, because both missed detections and false detections can terminate an otherwise effective exploration process. Recent methods therefore introduce verification or fusion mechanisms to improve grounding reliability. SG-Nav~\cite{26}  uses graph-based re-perception to reject low-credibility target candidates, and UniGoal~\cite{yin2025unigoal}  applies scene-graph correction and goal verification after graph matching. ApexNav~\cite{zhang2025apexnav} instead performs target-centric semantic fusion, aggregating multi-frame observations and confidence updates to suppress noisy false positives. These mechanisms mainly reduce false positives after a candidate has been detected, but do not recover targets that are visible yet missed by the detector. DRIVE-Nav addresses both cases: Qwen3-VL-guided prompt enrichment helps ground such missed targets with SAM3, while cross-frame verification filters false detections before map insertion.

\begin{figure*}[t]
\centering
 \includegraphics[width=0.84\textwidth,keepaspectratio]{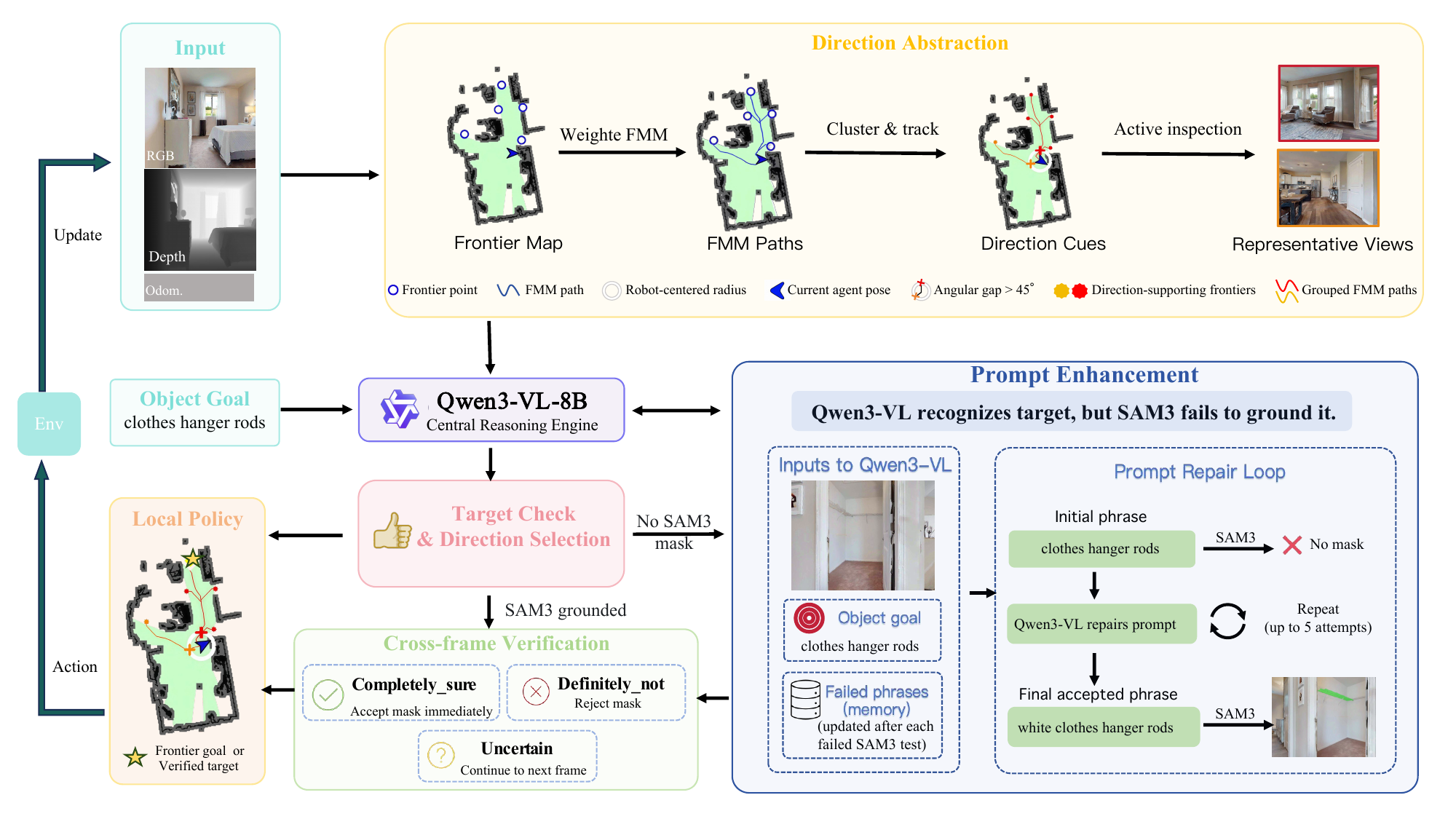}
\caption{Overview of the proposed closed-loop navigation pipeline. RGB-D observations are converted into persistent exits for direction selection, prompt enhancement, and cross-frame target verification. The robot-centered radius used for bearing extraction is defined in Sec.~IV-A(b).}
\label{fig:method_overview}
\end{figure*}
\section{Problem formulation}
We consider zero-shot open-vocabulary object navigation in an unseen
environment. At the beginning of each episode, the agent starts from an
initial pose $p_0$ and is given a language-specified target category
$g \in \mathcal{N}$. At each time step $t$, the agent receives an
egocentric RGB-D observation together with its estimated pose $p_t$ and
predicts an action. The goal is to navigate to an instance of category
$g$ in a previously unseen scene without task-specific training or
fine-tuning on the target categories. An episode is considered
successful if the agent reaches a target instance within the predefined
success distance and issues STOP before the navigation budget is
exhausted.

\section{Method}
\label{sec:method}

The overview of DRIVE-Nav is shown in \figref{fig:method_overview}.  From RGB-D observations and odometry, the agent  maintains a frontier map and converts raw frontier waypoints into persistent directions through weighted FMM  path analysis and angular clustering (Sec.~IV-A).  At episode start, the agent performs an initial $360^\circ$ scan, acquires representative direction views, and queries Qwen3-VL for  the first exploration direction. Thereafter, the low-level planner follows the farthest frontier point in the currently selected direction until an inspection event occurs: when the evolving frontier layout exposes newly inspectable directions within the forward $240^\circ$ range, the agent rotates to capture the missing representative views and Qwen3-VL selects the next direction (Sec.~IV-A(c)). If Qwen3-VL reports target visibility but SAM3 has not yet produced an accepted grounding, prompt enrichment converts direction-level evidence into detector-oriented SAM3 queries (Sec.~IV-B).  Accepted detections are further checked by  cross-frame verification  before being written into the object map (Sec.~IV-C).  High-level reasoning is thus triggered at direction-level inspection events or target-grounding events, while ordinary steps continue with frontier following in the active direction.

\subsection{Directional Reasoning over Persistent Directions}

We formulate exploration as directional reasoning over persistent directions, which convert raw frontier waypoints into stable directional representations and acquire representative views for inspectable directions within the forward $240^\circ$ range for semantic exploration.

\paragraph{ Path-induced direction candidates}
At each mapping step,  raw frontier waypoints are obtained from depth observations  using a standard frontier-map pipeline~\cite{yamauchi1997frontier,27}. Instead of grouping  these points directly by Euclidean position, we derive candidate directions from  weighted paths computed by the Fast Marching Method (FMM)~\cite{sethian1996fast}. We solve the Eikonal equation  $\|\nabla T(x)\|F(x)=1$ with $T(\mathbf{x}_r)=0$ at the robot position, using a spatially varying  speed field
\begin{equation}
F(x)=F_{\text{obs}}(x)F_{\text{vor}}(x),
\end{equation}
where  $F_{\text{obs}}$ and $F_{\text{vor}}$ encode obstacle avoidance and  medial-skeleton preference, adapted from distance-based FMM planning in VoroNav~\cite{wu2024voronav}:
\begin{equation}
F_{\text{obs}}(x)=1-\lambda \exp(-d_{\text{obs}}(x)/r_{\text{obs}}),
\end{equation}
 \begin{equation}
F_{\text{vor}}(x)=1+\beta \exp(-d_{\text{vor}}(x)/r_{\text{vor}}),
\end{equation}
with $d_{\text{obs}}$ and $d_{\text{vor}}$ the distances to the nearest obstacle  and  medial skeleton, respectively; $\lambda=0.65$, $r_{\text{obs}}=0.8$~m, $\beta=3.0$, and $r_{\text{vor}}=0.5$~m. For each frontier, the path is recovered by  gradient backtracking; the resulting reachable paths are then converted into local bearings in the next step.

\begin{figure*}[t]
\centering
\includegraphics[width=0.80\textwidth,keepaspectratio]{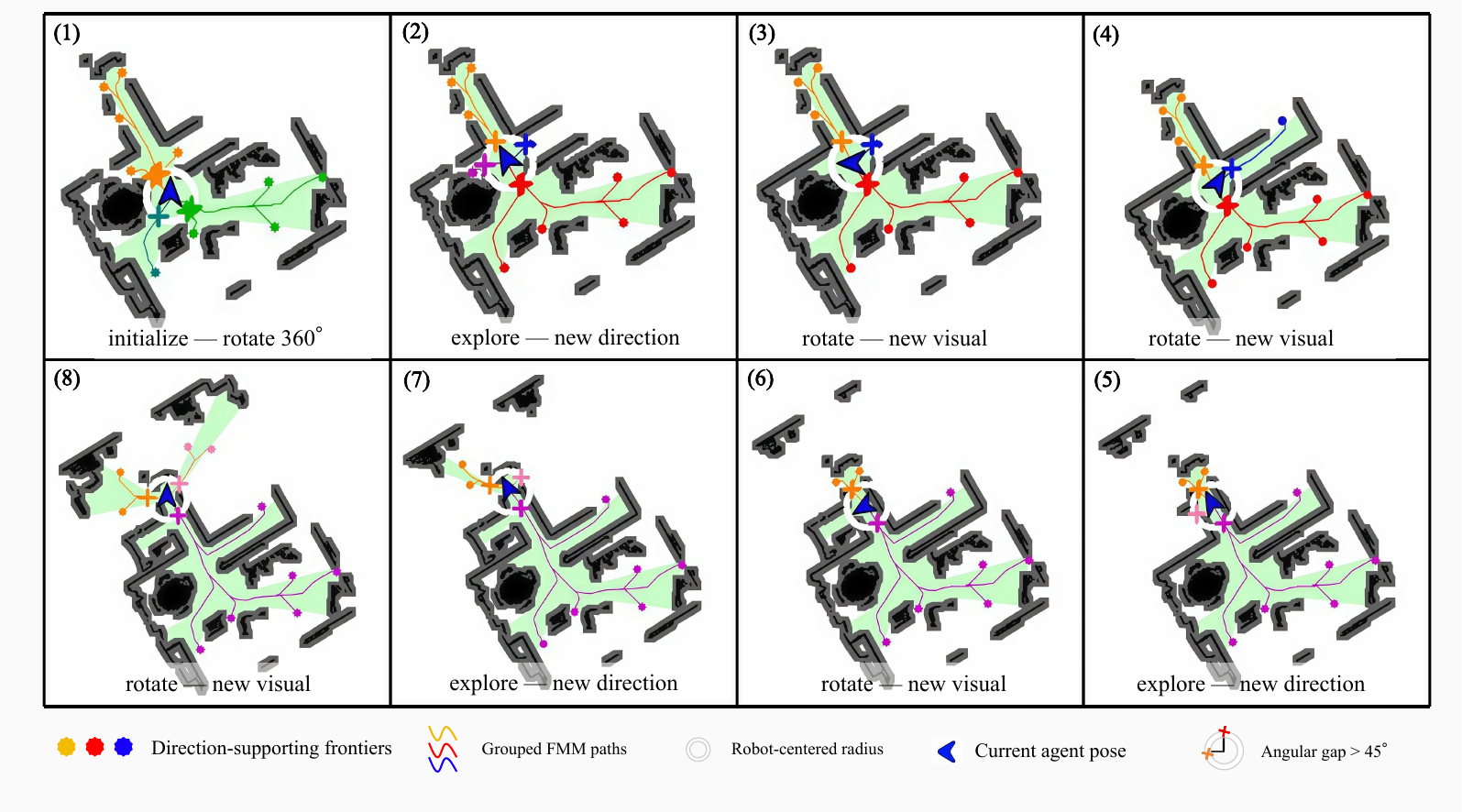}
\caption{Directional abstraction during a representative exploration episode. As the agent moves, local path-induced bearings are re-clustered into direction-supporting frontier sets shown by color. Panels (4)--(5) illustrate a dynamic update in which previously separate exits merge and a newly revealed bearing forms an additional direction.}
\label{fig:exit_abstraction}
\end{figure*}

\paragraph{Direction clustering and temporal association}
 We next convert each recovered FMM path into a local bearing measured in the episode world frame anchored at the robot's initial pose. For frontier $f_i$, we define its local directional point $\mathbf{p}_i$ as the first intersection between the FMM path and a robot-centered circle of radius $R_e=0.8$~m. This point is not a navigation goal, but only a geometric carrier for computing the path-induced bearing $\theta_i$ from the robot position $\mathbf{r}_t$ to $\mathbf{p}_i$. Each mapping step, the $\theta_i$ are normalized to $[0,2\pi)$, sorted circularly, and clustered by angular gaps: starting after the largest gap, bearings are greedily grouped while the span remains below $\tau_\theta=45^\circ$. This replaces Euclidean frontier grouping and handles the $0^\circ/360^\circ$ boundary. From \figref{fig:exit_abstraction}(4) to \figref{fig:exit_abstraction}(5), the blue and red exits change from being separated by more than $45^\circ$ to falling within the threshold and are therefore grouped into one direction, while the yellow direction reveals an additional bearing that forms a new pink direction. Each current-frame direction is
\begin{equation}
D_k^t = (\ell_k,\bar{\theta}_k^t),
\end{equation}
where $\bar{\theta}_k^t$ is the circular mean of member bearings and $\ell_k$ is a persistent label assigned at first appearance.

 Because clustering is recomputed as the map grows, temporal association maintains stable direction labels before view acquisition is performed. It has two stages. (1)~Position-based merge: if a new cluster's frontier waypoints co-occur at the same world-frame locations as those from two or more prior directions, the prior labels are removed and the merged cluster receives a fresh $\ell_k$. (2)~Angular association: for the remaining directions, we match representative bearings by circular distance
\begin{equation}
\Delta(\theta_a,\theta_b)=\min(|\theta_a-\theta_b|,\,2\pi-|\theta_a-\theta_b|),
\end{equation}
rank pairs by increasing $\Delta$, and greedily match unused ones; matches retain $\ell_k$ and update $\bar{\theta}_k^t$, unmatched current directions receive new $\ell_k$, and unmatched previous ones are removed. The resulting tracked directions provide the identities used by the representative-view module.

\paragraph{Tracked-direction view acquisition and semantic selection.}
Representative-view acquisition operates on the tracked directions from Sec.~IV-A(b). After the initial $360^\circ$ scan, the agent follows the farthest frontier in the selected direction until new branch directions become inspectable, then interrupts to acquire representative views for these newly revealed directions within the forward $240^\circ$ range. At an inspection event, we measure how far each representative bearing $\bar{\theta}_k^t$ deviates from the current robot heading $\psi_t$. Denote this signed turn angle by $\alpha_k=(\bar{\theta}_k^t-\psi_t)$, mapped to $(-\pi,\pi]$ so that $|\alpha_k|$ is the smallest rotation needed to face direction~$k$. Directions with $|\alpha_k|\le120^\circ$ form the forward $240^\circ$ inspectable set.
As shown in Fig. 3, directions in this set that lack valid representative views are arranged by signed offset and swept from the side requiring the smaller initial turn; directions with existing views do not trigger extra rotation, but their views are updated if the current yaw provides a smaller angular error. A representative view for $D_k^t$ is recorded once the camera yaw differs from $\bar{\theta}_k^t$ by less than $15^\circ$. Qwen3-VL then performs semantic selection over the acquired representative views, and the  low-level planner continues toward the farthest frontier point in the selected direction.

\subsection{Direction-Guided Prompt Enrichment}

 During navigation, SAM3 continuously attempts to segment the Habitat target category in the current RGB frame. Direction-level semantic inspection (Sec.~IV-A(c)) plays a complementary role: after representative views are acquired, Qwen3-VL selects the next exploration direction and checks whether the target  is visible. If no target is reported, the agent follows the selected direction . If the target is visible, Qwen3-VL returns the target-bearing direction  , and the robot turns toward that direction  for closer inspection.

 Prompt enrichment is activated only when Qwen3-VL has identified the target semantically but SAM3 has not yet returned an accepted mask. The module alternates between prompt proposal and grounding for up to five rounds: Qwen3-VL inspects the target-facing RGB view, the stored direction-level description, and previously failed prompts, then proposes a concise detector-oriented noun phrase; SAM3 immediately tests it on the current frame. An accepted mask initiates navigation toward the target, and the successful phrase is cached for later SAM3 calls.

\subsection{Cross-frame Verification}

To  improve grounding reliability, we  verify SAM3  target candidates with Qwen3-VL over at most three target-approach frames  collected during normal navigation. For each frame, Qwen3-VL receives the RGB image, a mask overlay, and previous verification frames, and outputs one of three judgments: \emph{ completely sure}, \emph{uncertain}, or \emph{ definitely not}.  A \emph{completely sure} judgment immediately accepts the candidate, \emph{uncertain} requests another approach frame, and \emph{definitely not} immediately rejects it. If no  rejection occurs within three frames, the candidate is accepted. Accepted masks are projected from RGB-D observations into the  object map  and can replace the frontier goal, whereas rejected detections are removed and stored as failed-position memory.

 \section{Experiments}
\label{sec:experiments}

\subsection{Experimental Setup}

\textbf{Datasets.} We evaluate our method in the Habitat simulator on  four benchmarks. HM3D-OVON serves as the primary benchmark for open-vocabulary evaluation , while HM3Dv1, HM3Dv2 and MP3D  provide additional closed-set ObjectNav  comparisons.

\textbf{Evaluation Metrics.} We report two standard ObjectNav metrics: Success Rate (SR) and Success weighted by Path Length (SPL). For the analytical study (Sec.~V-C), we additionally report  average executed steps and a revisit statistic to quantify action overhead and redundant re-exploration beyond what SPL captures.

\textbf{Implementation Details.} All experiments are conducted in Habitat with RGB-D observations and a maximum episode length of 500 steps. Instead of using a separate object detector, we directly adopt SAM3 as an open-vocabulary phrase segmentation model to produce target masks. For vision-language reasoning, including exit analysis, prompt enrichment, and semantic verification, we use Qwen3-VL-8B  served in bfloat16 with a maximum model length of 30,720 tokens and at most 30 images per prompt (video disabled). Main results use SAM3; grounding-backbone sensitivity in Sec.~V-D additionally evaluates OWLv2~\cite{minderer2023owlv2} with MobileSAM~\cite{zhang2023mobilesam} under the same directional-reasoning configuration without prompt enrichment or verification. All Habitat simulation experiments are conducted on a workstation equipped with four NVIDIA RTX 4090D GPUs.

\subsection{Comparison with State-of-the-art}

\begin{table*}[!t]
\centering
\caption{Comparison with state-of-the-art methods on HM3D-OVON, HM3Dv1, HM3Dv2, and MP3D. Zero-shot indicates no task-specific training on navigation data. The Detection, Segmentation, and Reasoning Model columns report the models used for target grounding, mask generation, and high-level reasoning, respectively, according to the corresponding method papers cited in the Method column. G-DINO denotes Grounding DINO~\cite{liu2024groundingdino}; ``--'' denotes not used. VoroNav reports HM3Dv1 only. Dark-blue/light-blue cells mark best/second-best per column.}
\label{tab:comparison_sr_spl}
\scriptsize
\setlength{\tabcolsep}{2.5pt}
\renewcommand{\arraystretch}{1.02}
\resizebox{\dimexpr\textwidth-6pt\relax}{!}{%
\begin{tabular}{@{}lcccccc *{8}{c}@{}}
\toprule
\multirow{2}{*}{Method} & \multirow{2}{*}{Venue} & \multirow{2}{*}{Zero-shot} & \multirow{2}{*}{Detection} & \multirow{2}{*}{Segmentation} & \multirow{2}{*}{Reasoning Model} & \multicolumn{2}{c}{HM3D-OVON} & \multicolumn{2}{c}{HM3Dv1} & \multicolumn{2}{c}{HM3Dv2} & \multicolumn{2}{c}{MP3D} \\
\cmidrule(lr){7-8} \cmidrule(lr){9-10} \cmidrule(lr){11-12} \cmidrule(lr){13-14}
& & & & & & SR$\uparrow$ & SPL$\uparrow$ & SR$\uparrow$ & SPL$\uparrow$ & SR$\uparrow$ & SPL$\uparrow$ & SR$\uparrow$ & SPL$\uparrow$ \\
\midrule
SemExp~\cite{5}   & CVPR'20 & $\times$ & Mask~R-CNN & Mask~R-CNN & -- & --   & --   & --   & --   & --   & --   & 36.0 & 14.4 \\
PONI~\cite{19}    & CVPR'22    & $\times$ & -- & RedNet & -- & --   & --   & --   & --   & --   & --   & 31.8 & 12.1 \\
ZSON~\cite{15}    & NeurIPS'22 & $\times$ & -- & -- & -- & --   & --   & 25.5 & 12.6 & --   & --   & 15.3 & 4.8  \\
ActPept~\cite{guo2024actpept} & RAL'24 & $\times$ & -- & -- & GCN & --   & --   & --   & --   & --   & --   & 39.8 & 17.4 \\
\midrule
CoW~\cite{7}           & CVPR'23    & $\checkmark$ & OWL-ViT & -- & -- & --   & --   & --   & --   & --   & --   & 7.4  & 3.7  \\
ESC~\cite{32}          & ICML'23    & $\checkmark$ & GLIP-L & -- & DeBERTa-v3 & --   & --   & 39.2 & 22.3 & --   & --   & 28.7 & 14.2 \\
L3MVN~\cite{28}        & IROS'23    & $\checkmark$ & -- & RedNet & RoBERTa-large & --   & --   & 50.4 & 23.1 & 36.3 & 15.7 & --   & --   \\
OpenFMNav~\cite{kuang2024openfmnav} & NAACL-F'24 & $\checkmark$ & G-DINO & SAM & GPT-4V & --   & --   & 54.9 & 24.4 & --   & --   & 37.2 & 15.7 \\
InstructNav~\cite{14}  & CoRL'24   & $\checkmark$ & GLEE & GLEE & GPT-4V & --   & --   & --   & --   & 58.0 & 20.9 & --   & --   \\
SG-Nav~\cite{26}       & NeurIPS'24 & $\checkmark$ & GLIP/G-DINO & SAM & GPT-4 & --   & --   & 54.0 & 24.9 & 49.6 & 25.5 & 40.2 & 16.0 \\
UniGoal~\cite{yin2025unigoal}  & CVPR'25  & $\checkmark$ & GLIP/G-DINO & SAM & GPT-4 & --   & --   & 54.5  & 25.1   & --   & --   & \secondcell{41.0} & 16.4 \\
ApexNav~\cite{zhang2025apexnav} & RAL'25 & $\checkmark$ & YOLOv7+G-DINO & MobileSAM & -- & --   & --   & \bestcell{59.6} & \secondcell{33.0} & \bestcell{76.2} & \secondcell{38.0} & 39.2 & \secondcell{17.8} \\
VoroNav~\cite{wu2024voronav} & ICML'24 & $\checkmark$ & G-DINO & SAM & GPT-3.5 & --   & --   & 42.0 & 26.0 & --   & --   & --   & --   \\
VLFM~\cite{27,40}    & ICRA'24    & $\checkmark$ & OWLv2/G-DINO & MobileSAM & -- & 35.2 & 19.6 & 52.5 & 30.4 & 63.6 & 32.5 & 36.4 & 17.5 \\
MSGNav~\cite{huang2025msgnav} & CVPR'26 & $\checkmark$ & YOLO-World & SAM & GPT-4o & \secondcell{48.3} & \secondcell{27.0} & --   & --   & --   & --   & --   & --   \\
\midrule
\textbf{DRIVE-Nav (Ours)} & \textbf{Ours} & $\checkmark$ & SAM3 & SAM3 & Qwen3-VL-8B & \bestcell{50.2} & \bestcell{32.6} & \secondcell{55.5} & \bestcell{35.9} & \secondcell{72.4} & \bestcell{41.3} & \bestcell{41.8} & \bestcell{22.6} \\
\bottomrule
\end{tabular}%
}
\end{table*}

\tabref{tab:comparison_sr_spl} compares DRIVE-Nav with state-of-the-art object navigation methods on HM3D-OVON, HM3Dv1, HM3Dv2, and MP3D. Blank cells indicate that the corresponding method was not evaluated on a given benchmark. Backbone differences are summarized in the Detection, Segmentation, and Reasoning Model columns, while matched module and grounding-backbone ablations are provided in Sec.~V-D.

On the primary HM3D-OVON benchmark, DRIVE-Nav achieves the best overall result (50.2\% SR; 32.6\% SPL), outperforming MSGNav~\cite{huang2025msgnav} by 1.9 points in SR and 5.6 points in SPL. On HM3Dv1, DRIVE-Nav achieves 55.5\% SR and 35.9\% SPL, improving over the closest topology-aware baseline VoroNav~\cite{wu2024voronav} (42.0\%/26.0\%) by 13.5 SR and 9.9 SPL points. On HM3Dv2, it achieves the best SPL of 41.3\%, exceeding ApexNav~\cite{zhang2025apexnav} and VLFM~\cite{27} by 3.3 and 8.8 points, respectively, while maintaining 72.4\% SR. On MP3D, DRIVE-Nav again delivers the best overall result (41.8\% SR; 22.6\% SPL), improving the best prior SPL from ApexNav~\cite{zhang2025apexnav} by 4.8 points and from SG-Nav~\cite{26} by 6.6 points. These consistent SPL gains validate the efficiency of our directional-reasoning and inspection-driven navigation strategy.

\FloatBarrier
\subsection{Analytical Study}
\label{subsec:analytical_study}

 Standard SPL normalizes path length by geodesic distance, but it does not reveal action waste from in-place rotations  or repeated traversal. We therefore compare episodes with start-to-goal distance above 10\,m that are successfully completed by all methods.

We compare VLFM, a nearest-edge-point greedy policy,  a direction-aware  policy with full $360^\circ$ inspection, and DRIVE-Nav. \emph{Average steps} counts all executed actions, including rotations. \emph{Average revisits} counts a frontier-selection segment as a revisit if it accumulates at least 5\,m of travel within 0.5\,m of the previously traversed 2D trajectory; this threshold avoids counting brief overlaps caused by normal room entry and exit.

 \begin{table}[!t]
\centering
\caption{Analytical comparison  on common successful episodes with start-to-goal distance $>$ 10\,m . Lower is better.}
\label{tab:step_analysis}
\footnotesize
 \begin{tabular}{lcc}
\toprule
Method & Average steps & Avg. revisits \\
\midrule
VLFM~\cite{27} & 163.5 & 0.364 \\
Nearest-edge-point greedy & 164.8 & 0.200 \\
Direction inspection + 360$^\circ$ scan & 222.1 & 0.200 \\
\textbf{DRIVE-Nav (Ours)} & \textbf{133.4} & \textbf{0.132} \\
\bottomrule
\end{tabular}
\end{table}

 As shown in \tabref{tab:step_analysis},  DRIVE-Nav requires the fewest steps  (133.4 ) and revisits (0.132). Compared with VLFM and the full $360^\circ$ inspection baseline, the results show that  direction tracking and forward-range inspection reduce both rotation overhead and redundant re-exploration.

 \FloatBarrier
\begin{figure*}[!b]
\centering
\includegraphics[width=0.90\textwidth,keepaspectratio]{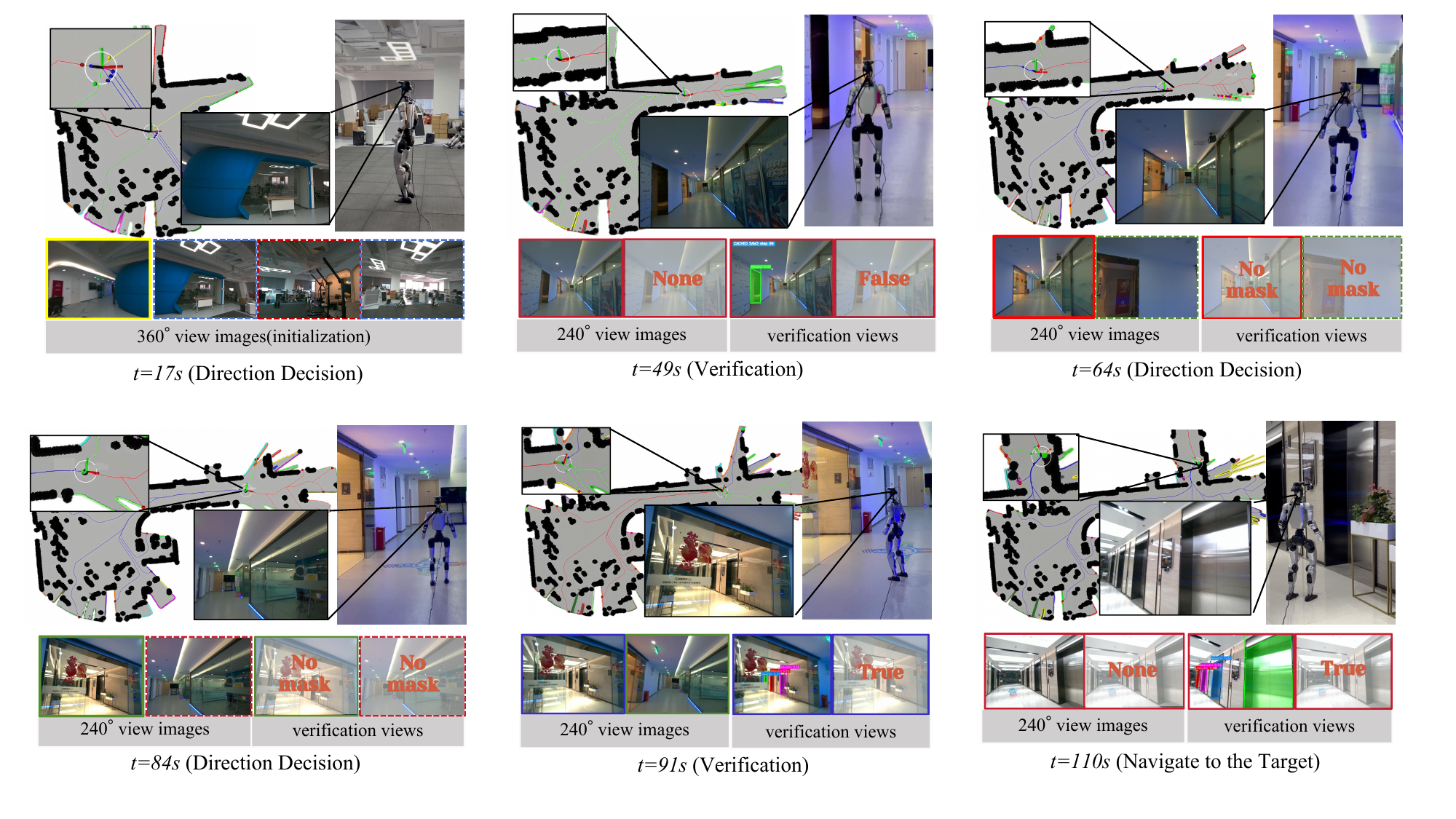}
\caption{Real-world deployment of DRIVE-Nav for finding an elevator. The sequence shows the full navigation episode, including $360^\circ$ initialization, successive $240^\circ$ direction decisions, cross-frame verification, and final approach to the target, with colored boxes indicating candidate directions in the same color and solid lines marking the selected direction.}
\label{fig:real_world}
\end{figure*}

\subsection{Ablation Study}
\label{subsec:ablation}

We ablate DRIVE-Nav on the full HM3D-OVON validation set (3000 episodes) to isolate directional components, grounding-backbone effects, and the final semantic modules.

\textbf{Directional reasoning components.} \tabref{tab:ablation_geometry} shows that weighted FMM and $240^\circ$ inspection are complementary: removing either component lowers SR/SPL, with weighted FMM causing the larger SR drop and full $360^\circ$ inspection increasing unnecessary rotations.

\begin{table}[!t]
\centering
\caption{ Directional-reasoning component ablation on HM3D-OVON. }
\label{tab:ablation_geometry}
\footnotesize
\begin{tabular}{lcc}
\toprule
Method & SR$\uparrow$ & SPL$\uparrow$ \\
\midrule
w/o Weighted FMM & 44.5 & 26.2 \\
w/o $240^\circ$ Inspection View & 45.5 & 26.8 \\
\textbf{Full Directional Reasoning Pipeline} & \textbf{46.3} & \textbf{27.5} \\
\bottomrule
\end{tabular}
\end{table}

 \textbf{Grounding-backbone sensitivity.} Under the directional-reasoning-only setting, replacing OWLv2+MobileSAM with SAM3 yields only a modest gain (+0.7 SR; +1.2 SPL), indicating that the full-system gain mainly comes from prompt enrichment and verification.

 \begin{table}[!t]
\centering
\caption{Grounding comparison under matched modules on HM3D-OVON.}
\label{tab:ablation_grounding}
\footnotesize
\begin{tabular}{lcc}
\toprule
Grounding & SR$\uparrow$ & SPL$\uparrow$ \\
\midrule
OWLv2~\cite{minderer2023owlv2}+MobileSAM~\cite{zhang2023mobilesam} & 45.6 & 26.3 \\
SAM3~\cite{carion2025sam} & \textbf{46.3} & \textbf{27.5} \\
\bottomrule
\end{tabular}
\end{table}

 \textbf{ Module ablation.} \tabref{tab:ablation_semantic}  uses fixed SAM3 and Qwen3-VL-8B backbones. Directional reasoning improves the frontier baseline from 42.1\%/22.4\% to 46.3\%/27.5\%, and prompt enrichment plus verification further raise the full system to 50.2\% /32.6\% .

 \begin{table}[!t]
\centering
\caption{ Module ablation on HM3D-OVON with fixed SAM3 and Qwen3-VL-8B backbones. Higher is better.}
\label{tab:ablation_semantic}
\footnotesize
\begin{tabular*}{\columnwidth}{@{\extracolsep{\fill}}cccc cc@{}}
\toprule
\multicolumn{4}{c}{Module} & \multirow{2}{*}[-3pt]{SR$\uparrow$} & \multirow{2}{*}[-3pt]{SPL$\uparrow$} \\
\cmidrule(r){1-4}
Frontier & Directional & Prompt Enrich. & Verification &  &  \\
\midrule
$\checkmark$ &  &  &  & 42.1 & 22.4 \\
$\checkmark$ & $\checkmark$ &  &  & 46.3 & 27.5 \\
$\checkmark$ & $\checkmark$ & $\checkmark$ &  & 48.0 & 28.9 \\
$\checkmark$ & $\checkmark$ &  & $\checkmark$ & 48.8 & 30.8 \\
$\checkmark$ & $\checkmark$ & $\checkmark$ & $\checkmark$ & \textbf{50.2} & \textbf{32.6} \\
\bottomrule
\end{tabular*}
\end{table}

 \subsection{Real-world Deployment}
\label{subsec:real_world}

DRIVE-Nav is deployed on a Unitree G1 humanoid robot  with onboard sensing and offboard computation.  A RealSense D455  on a head-mounted two-axis  gimbal provides forward RGB-D observations , while LiDAR and IMU data are  fused by FAST-LIO2~\cite{xu2022fastlio2}  and projected into a 2D occupancy grid . To match the Habitat sensing setup, where the RGB-D camera has a horizontal FOV of $79^\circ$, only LiDAR returns within  the same forward-facing $79^\circ$  range centered on the camera are used  for occupancy projection; raw frontier waypoints are then extracted and converted into persistent directions. The high-level pipeline runs on a single RTX 4090 through ROS2-ZMQ; Qwen3-VL-4B  is used instead of the 8B simulation model so that direction selection, prompt enrichment, verification, and SAM3  grounding can run on the same GPU.

\textbf{ Computational analysis.}  \tabref{tab:runtime} profiles the deployment workstation. SAM3 dominates synchronous per-step cost ($\sim$270\,ms), while computing the FMM path-induced direction for one frontier point takes $\sim$15\,ms on CPU. Qwen3-VL-4B is invoked only at sparse decision points, so the deployed system is perception-limited rather than reasoning-limited.

\begin{table}[!htbp]
\centering
\caption{Runtime on the real-world workstation.}
\label{tab:runtime}
\footnotesize
\setlength{\tabcolsep}{4pt}
\begin{tabular*}{\columnwidth}{@{\extracolsep{\fill}}lll@{}}
\toprule
Module & Latency & Invocation \\
\midrule
SAM3 (detection + semantic map) & $\sim$270\,ms & every step \\
FMM candidate generation & $\sim$15\,ms & per solve \\
Direction selection & $\sim$8\,s & few per episode \\
Prompt enrichment & $\sim$3\,s & optional, sparse \\
Cross-frame verification & $\sim$3\,s & up to 3 frames \\
\bottomrule
\end{tabular*}
\end{table}

\textbf{Experimental results.}
Representative real-world demonstrations cover five target categories: \emph{toilet}, \emph{elevator}, \emph{plant}, \emph{fire extinguisher}, and \emph{vending machine}. In these demonstrations, the robot successfully locates and reaches each target. We also observe that real LiDAR-based boundary estimation may require short sensing stabilization, which can introduce transient frontier changes and temporarily trigger direction reconsideration. Larger-scale real-robot benchmarking and baseline deployment remain future work. \figref{fig:real_world} shows a representative elevator trial.

\FloatBarrier
\section{Conclusion}
This paper presented DRIVE-Nav, a zero-shot open-vocabulary navigation framework that unifies directional reasoning, inspection, and verification in a single decision loop. By organizing exploration around persistent directions rather than dense frontier points, DRIVE-Nav enables more stable route selection, more efficient inspection, and more reliable target grounding through prompt enrichment and cross-frame verification. Experiments on HM3D-OVON, HM3Dv1, HM3Dv2, and MP3D show consistent improvements in overall navigation performance, including strong gains in SPL across all  four benchmarks. The analytical study further indicates that the actual efficiency advantage of DRIVE-Nav is more substantial than suggested by SPL alone.  Proof-of-concept deployment on a physical humanoid robot  suggests practical transfer beyond simulation . Future work will extend DRIVE-Nav to  dynamic scenes, larger-scale real-robot benchmarking, and episodic memory for revisitation-aware planning.

\bibliographystyle{IEEEtran}
\bibliography{reference}

@article{zhang2025apexnav,
  author = {Zhang, Mingjie and Du, Yuheng and Wu, Chengkai and Zhou, Jinni and Qi, Zhenchao and Ma, Jun and Zhou, Boyu},
  title = {{ApexNav}: An adaptive exploration strategy for zero-shot object navigation with target-centric semantic fusion},
  journal = {IEEE Robot. Autom. Lett.},
  year = {2025},
  publisher = {IEEE}
}

@article{wu2024voronav,
  author = {Wu, Pengying and Mu, Yao and Wu, Bingxian and Hou, Yi and Ma, Ji and Zhang, Shanghang and Liu, Chang},
  title = {{VoroNav}: Voronoi-based zero-shot object navigation with large language model},
  journal = {arXiv preprint arXiv:2401.02695},
  year = {2024}
}

@article{huang2025msgnav,
  author = {Huang, Xun and Zhao, Shijia and Wang, Yunxiang and Lu, Xin and Zhang, Wanfa and Qu, Rongsheng and Li, Weixin and Wang, Yunhong and Wen, Chenglu},
  title = {{MSGNav}: Unleashing the power of multi-modal {3D} scene graph for zero-shot embodied navigation},
  journal = {arXiv preprint arXiv:2511.10376},
  year = {2025}
}

@article{5,
  author = {Chaplot, Devendra Singh and Gandhi, Dhiraj Prakashchand and Gupta, Abhinav and Salakhutdinov, Russ R},
  title = {Object goal navigation using goal-oriented semantic exploration},
  journal = {Advances in Neural Information Processing Systems},
  volume = {33},
  pages = {4247--4258},
  year = {2020}
}

@inproceedings{7,
  author = {Gadre, Samir Yitzhak and Wortsman, Mitchell and Ilharco, Gabriel and Schmidt, Ludwig and Song, Shuran},
  title = {Cows on pasture: Baselines and benchmarks for language-driven zero-shot object navigation},
  booktitle = {Proceedings of the IEEE/CVF Conference on Computer Vision and Pattern Recognition},
  pages = {23171--23181},
  year = {2023}
}

@article{14,
  author = {Long, Yuxing and Cai, Wenzhe and Wang, Hongcheng and Zhan, Guanqi and Dong, Hao},
  title = {{InstructNav}: Zero-shot system for generic instruction navigation in unexplored environment},
  journal = {arXiv preprint arXiv:2406.04882},
  year = {2024}
}

@article{15,
  author = {Majumdar, Arjun and Aggarwal, Gunjan and Devnani, Bhavika and Hoffman, Judy and Batra, Dhruv},
  title = {{ZSON}: Zero-shot object-goal navigation using multimodal goal embeddings},
  journal = {Advances in Neural Information Processing Systems},
  volume = {35},
  pages = {32340--32352},
  year = {2022}
}

@inproceedings{19,
  author = {Ramakrishnan, Santhosh Kumar and others},
  title = {{PONI}: Potential functions for {ObjectGoal} navigation with interaction-free learning},
  booktitle = {Proceedings of the IEEE/CVF Conference on Computer Vision and Pattern Recognition},
  year = {2022}
}

@inproceedings{26,
  author = {Lin, Bingqian and others},
  title = {{SG-Nav}: Online {3D} scene graph prompting for {LLM}-based zero-shot object navigation},
  booktitle = {Advances in Neural Information Processing Systems},
  year = {2024}
}

@inproceedings{27,
  author = {Yokoyama, Naoki and Ha, Sehoon and Batra, Dhruv and Wang, Jiuguang and Bucher, Bernadette},
  title = {{VLFM}: Vision-language frontier maps for zero-shot semantic navigation},
  booktitle = {2024 IEEE International Conference on Robotics and Automation (ICRA)},
  pages = {42--48},
  year = {2024},
  organization = {IEEE}
}

@inproceedings{28,
  author = {Yu, Bangguo and Kasaei, Hamidreza and Cao, Ming},
  title = {{L3MVN}: Leveraging large language models for visual target navigation},
  booktitle = {2023 IEEE/RSJ International Conference on Intelligent Robots and Systems (IROS)},
  pages = {3554--3560},
  year = {2023},
  organization = {IEEE}
}

@inproceedings{32,
  author = {Zhou, Kaiwen and Zheng, Kaizhi and Pryor, Connor and Shen, Yilin and Jin, Hongxia and Getoor, Lise and Wang, Xin Eric},
  title = {{ESC}: Exploration with soft commonsense constraints for zero-shot object navigation},
  booktitle = {International Conference on Machine Learning},
  pages = {42829--42842},
  year = {2023},
  organization = {PMLR}
}

@inproceedings{40,
  author = {Yokoyama, Naoki and Ramrakhya, Ram and Das, Abhishek and Batra, Dhruv and Ha, Sehoon},
  title = {{HM3D}-{OVON}: A dataset and benchmark for open-vocabulary object goal navigation},
  booktitle = {2024 IEEE/RSJ International Conference on Intelligent Robots and Systems (IROS)},
  pages = {5543--5550},
  year = {2024},
  organization = {IEEE}
}

@inproceedings{yamauchi1997frontier,
  author = {Yamauchi, Brian},
  title = {A frontier-based approach for autonomous exploration},
  booktitle = {Proceedings of the 1997 IEEE International Symposium on Computational Intelligence in Robotics and Automation},
  pages = {146--151},
  year = {1997},
  organization = {IEEE}
}

@article{batra2020objectnav,
  author = {Batra, Dhruv and Gokaslan, Aaron and Kembhavi, Aniruddha and Maksymets, Oleksandr and Mottaghi, Roozbeh and Savva, Manolis and Toshev, Alexander and Wijmans, Erik},
  title = {{ObjectNav} revisited: On evaluation of embodied agents navigating to objects},
  journal = {arXiv preprint arXiv:2006.13171},
  year = {2020}
}

@inproceedings{yin2025unigoal,
  author = {Yin, Hang and Xu, Xiuwei and Zhao, Linqing and Wang, Ziwei and Zhou, Jie and Lu, Jiwen},
  title = {{UniGoal}: Towards universal zero-shot goal-oriented navigation},
  booktitle = {Proceedings of the IEEE/CVF Conference on Computer Vision and Pattern Recognition},
  pages = {19057--19066},
  year = {2025}
}

@inproceedings{yadav2023habitat,
  author = {Yadav, Karmesh and Ramrakhya, Ram and Ramakrishnan, Santhosh Kumar and Gervet, Theo and Turner, John and Gokaslan, Aaron and Maestre, Noah and Chang, Angel Xuan and Batra, Dhruv and Savva, Manolis and others},
  title = {{Habitat}-{Matterport} {3D} semantics dataset},
  booktitle = {Proceedings of the IEEE/CVF Conference on Computer Vision and Pattern Recognition},
  pages = {4927--4936},
  year = {2023}
}

@article{ramakrishnan2021hm3d,
  author = {Ramakrishnan, Santhosh K. and Gokaslan, Aaron and Wijmans, Erik and Maksymets, Oleksandr and Clegg, Alexander and Turner, John and Undersander, Eric and Galuba, Wojciech and Westbury, Andrew and Chang, Angel X. and Savva, Manolis and Zhao, Yili and Batra, Dhruv},
  title = {{Habitat}-{Matterport} 3D dataset ({HM3D}): 1000 large-scale 3D environments for embodied AI},
  journal = {arXiv preprint arXiv:2109.08238},
  year = {2021}
}

@article{chang2017matterport3d,
  author = {Chang, Angel and Dai, Angela and Funkhouser, Thomas and Halber, Maciej and Niessner, Matthias and Savva, Manolis and Song, Shuran and Zeng, Andy and Zhang, Yinda},
  title = {{Matterport3D}: Learning from {RGB}-{D} data in indoor environments},
  journal = {arXiv preprint arXiv:1709.06158},
  year = {2017}
}

@article{xu2022fastlio2,
  author = {Xu, Wei and Cai, Yixi and He, Dongjiao and Lin, Jiarong and Zhang, Fu},
  title = {{FAST-LIO2}: Fast direct {LiDAR}-inertial odometry},
  journal = {IEEE Transactions on Robotics},
  volume = {38},
  number = {4},
  pages = {2053--2073},
  year = {2022},
  publisher = {IEEE}
}

@article{sethian1996fast,
  author = {Sethian, James A},
  title = {A fast marching level set method for monotonically advancing fronts},
  journal = {Proceedings of the National Academy of Sciences},
  volume = {93},
  number = {4},
  pages = {1591--1595},
  year = {1996}
}

@article{bai2025qwen3,
  author = {Bai, Shuai and Cai, Yuxuan and Chen, Ruizhe and Chen, Keqin and Chen, Xionghui and Cheng, Zesen and Deng, Lianghao and Ding, Wei and Gao, Chang and Ge, Chunjiang and others},
  title = {{Qwen3}-{VL} technical report},
  journal = {arXiv preprint arXiv:2511.21631},
  year = {2025}
}

@article{carion2025sam,
  author = {Carion, Nicolas and Gustafson, Laura and Hu, Yuan-Ting and Debnath, Shoubhik and Hu, Ronghang and Suris, Didac and Ryali, Chaitanya and Alwala, Kalyan Vasudev and Khedr, Haitham and Huang, Andrew and others},
  title = {{SAM} 3: Segment anything with concepts},
  journal = {arXiv preprint arXiv:2511.16719},
  year = {2025}
}

@article{kuang2024openfmnav,
  author = {Kuang, Y. and Lin, H. and Jiang, M.},
  title = {{OpenFMNav}: Towards open-set zero-shot object navigation via vision-language foundation models},
  journal = {arXiv preprint arXiv:2402.10670},
  year = {2024}
}

@article{guo2024actpept,
  author = {Guo, Y. and others},
  title = {An object-driven navigation strategy based on active perception and semantic association},
  journal = {IEEE Robotics and Automation Letters},
  volume = {9},
  number = {8},
  pages = {7110--7117},
  year = {2024}
}

@article{brunye2018spatial,
  author = {Bruny{\'e}, Tad T. and Gardony, Aaron L. and Holmes, Amanda and Taylor, Holly A.},
  title = {Spatial decision dynamics during wayfinding: Intersections prompt the decision-making process},
  journal = {Cognitive Research: Principles and Implications},
  volume = {3},
  number = {13},
  year = {2018},
  doi = {10.1186/s41235-018-0098-3}
}

@inproceedings{minderer2023owlv2,
  author = {Minderer, Matthias and Gritsenko, Alexey and Houlsby, Neil},
  title = {Scaling open-vocabulary object detection},
  booktitle = {Advances in Neural Information Processing Systems},
  year = {2023}
}

@article{zhang2023mobilesam,
  author = {Zhang, Chaoning and Han, Dongshen and Qiao, Yu and Kim, Jung Uk and Bae, Sung-Ho and Lee, Seungkyu and Hong, Choong Seon},
  title = {Faster segment anything: Towards lightweight {SAM} for mobile applications},
  journal = {arXiv preprint arXiv:2306.14289},
  year = {2023}
}

@inproceedings{liu2024groundingdino,
  author = {Liu, Shilong and Zeng, Zhaoyang and others},
  title = {Grounding {DINO}: Marrying {DINO} with Grounded Pre-Training for Open-Set Object Detection},
  booktitle = {European Conference on Computer Vision},
  pages = {38--55},
  publisher = {Springer},
  year = {2024}
}
\end{document}